\documentclass{article}

\usepackage{arxiv}
\usepackage[utf8]{inputenc}
\usepackage[T1]{fontenc}
\usepackage{lmodern}
\usepackage{microtype}
\usepackage{natbib}
\usepackage[hyphens]{url}
\usepackage{graphicx}
\usepackage{amsmath,amssymb}
\usepackage{booktabs}
\usepackage{tabularx}
\usepackage{threeparttable}
\usepackage{caption}
\usepackage{xspace}
\usepackage{xcolor}
\usepackage{hyperref}

\hypersetup{
  colorlinks=true,
  linkcolor=blue!40!black,
  citecolor=blue!40!black,
  urlcolor=blue!40!black,
  pdftitle={FemWear: A Parameter-Efficient Wearable Foundation Model for Women's Health},
  pdfauthor={Yifan Wang and Chenzhong Li},
  pdfkeywords={wearable foundation model, women's health, parameter-efficient fine-tuning, multitask learning, menstrual health}
}

\graphicspath{{figures/}}
\newcommand{\method}{\textsc{FemWear}\xspace}
\newcommand{\openmhc}{\textsc{OpenMHC}\xspace}
\newcommand{\better}[1]{\textbf{#1}}
\newcolumntype{Y}{>{\raggedright\arraybackslash}X}
\newcolumntype{R}{>{\raggedleft\arraybackslash}X}
\title{FemWear: A Parameter-Efficient Wearable Foundation Model for Women's Health}

\author{
  Yifan Wang\\
  The Chinese University of Hong Kong, Shenzhen\\
  \texttt{224050081@link.cuhk.edu.cn}
  \And
  Chenzhong Li\\
  The Chinese University of Hong Kong, Shenzhen\\
  \texttt{lichenzhong@cuhk.edu.cn}
}

\date{\today}
\renewcommand{\shorttitle}{FemWear: A Wearable Foundation Model for Women's Health}
\renewcommand{\headeright}{}
\renewcommand{\undertitle}{}

\begin{document}
\maketitle

\begin{abstract}
General-purpose wearable foundation models are pretrained on broad sensor streams and populations, but their representations are not organized around women's health. \method is a women's wearable foundation model, obtained by parameter-efficiently repurposing a pretrained general multimodal wearable backbone into a specialized representation for women's health. It keeps the pretrained patch projection and Transformer encoder frozen and trains 239,236 encoder parameters---1.11\% of a 21.54M-parameter encoder---through low-rank residual adapters and causal task-family heads, producing one shared longitudinal representation for menstrual, symptom, affective, sleep/recovery, autonomic, activity, and pregnancy outcomes. We evaluate six cohorts with 63 comparable primary metrics, 33 from women's-health cohorts, while retaining the 32-task \openmhc ability-retention benchmark. On a fixed participant split over three seeds, \method improved cycle-phase macro-F1 by 8.15\% and reduced mean absolute error for cramps, mood symptoms, and sleep problems by 9.32\%, 5.80\%, and 9.43\%; 24-hour onset AUPRC decreased by 3.40\%. A stricter 42-participant nested leave-one-participant-out audit retained positive changes for 24-hour onset (+2.87\%), 72-hour onset (+6.35\%), and cramps (+2.19\%), while phase, mood, and sleep changes were neutral or negative and no endpoint had a strictly positive corrected confidence interval. Capacity-matched experiments beat a latest-day multilayer perceptron but not shared-GRU or multi-gate mixture-of-experts baselines. Train-only calibration reduced onset expected calibration error by 84.2--88.2\% with zero temporal-nesting violations. \method is therefore a women's wearable foundation model: a reproducible, parameter-efficient specialization delivering targeted transfer across women's-health tasks and coherent probability outputs.
\end{abstract}


\section{Introduction}

Consumer wearables now measure activity, sleep, heart rate, heart-rate variability, and temperature continuously in everyday life. Recent wearable foundation models use self-supervised pretraining to transfer representations across sensor modalities and downstream tasks \citep{narayanswamy2024scaling,pillai2025papagei}. \openmhc extends this direction with open data, model implementations, and a 32-task benchmark built from more than 60 million hours of wearable data \citep{schuetz2026openmhc}. These resources make it practical to build specialized models on top of a general wearable representation; what they do not provide is a representation already organized around women's-health dynamics or outcomes.

Women's wearable health is a heterogeneous longitudinal modeling problem. Menstrual phase and onset vary over days; symptoms and affect are sparse and self-reported; sleep and autonomic targets use different observation windows; and pregnancy activity follows a distinct longitudinal trajectory. Physiological studies show that wrist temperature and nocturnal pulse vary across the menstrual cycle \citep{lin2024temperature,shilaih2017pulse}, and the mcPHASES dataset links these wearable signals to daily symptoms and urine hormone measurements \citep{lin2025mcphases}. The supervision needed to model these phenomena is scattered across small cohorts, different devices, and partially overlapping task families. Training one predictor per task discards shared longitudinal structure, while unrestricted multitask sharing can create negative transfer when objectives compete \citep{standley2020tasks}.

This work asks whether a general wearable foundation representation can be repurposed as a specialized model for women's health with a small trainable parameter budget, and where that specialization transfers. We introduce \method, a women's wearable foundation model that specializes the \openmhc backbone through low-rank residual adaptation, semantic sensor alignment, causal longitudinal aggregation, and task-family output heads. We evaluate it on menstrual, affective, sleep/HRV, pregnancy, and general ability-retention cohorts under participant-disjoint protocols. The experimental design deliberately measures positive transfer, neutral transfer, and degradation, rather than treating specialization as synonymous with improvement on every endpoint.

The paper contributes three things. First, \method itself: a shared representation and task-family interface spanning heterogeneous women's-health cohorts, built by training only 1.11\% of a general encoder. Second, a reproducible evaluation suite with 63 comparable primary metrics across six sources, including 33 from women's-health cohorts, that pits a 1.11\%-trainable specialization against static, recurrent, and mixture-of-experts baselines under matched data, parameter counts, seeds, and fixed training budgets. Third, a rigorous account of where specialization transfers, established through nested participant validation, calibration, label-efficiency, and missing-history audits.

\begin{figure}[t]
  \centering
  \includegraphics[width=\textwidth]{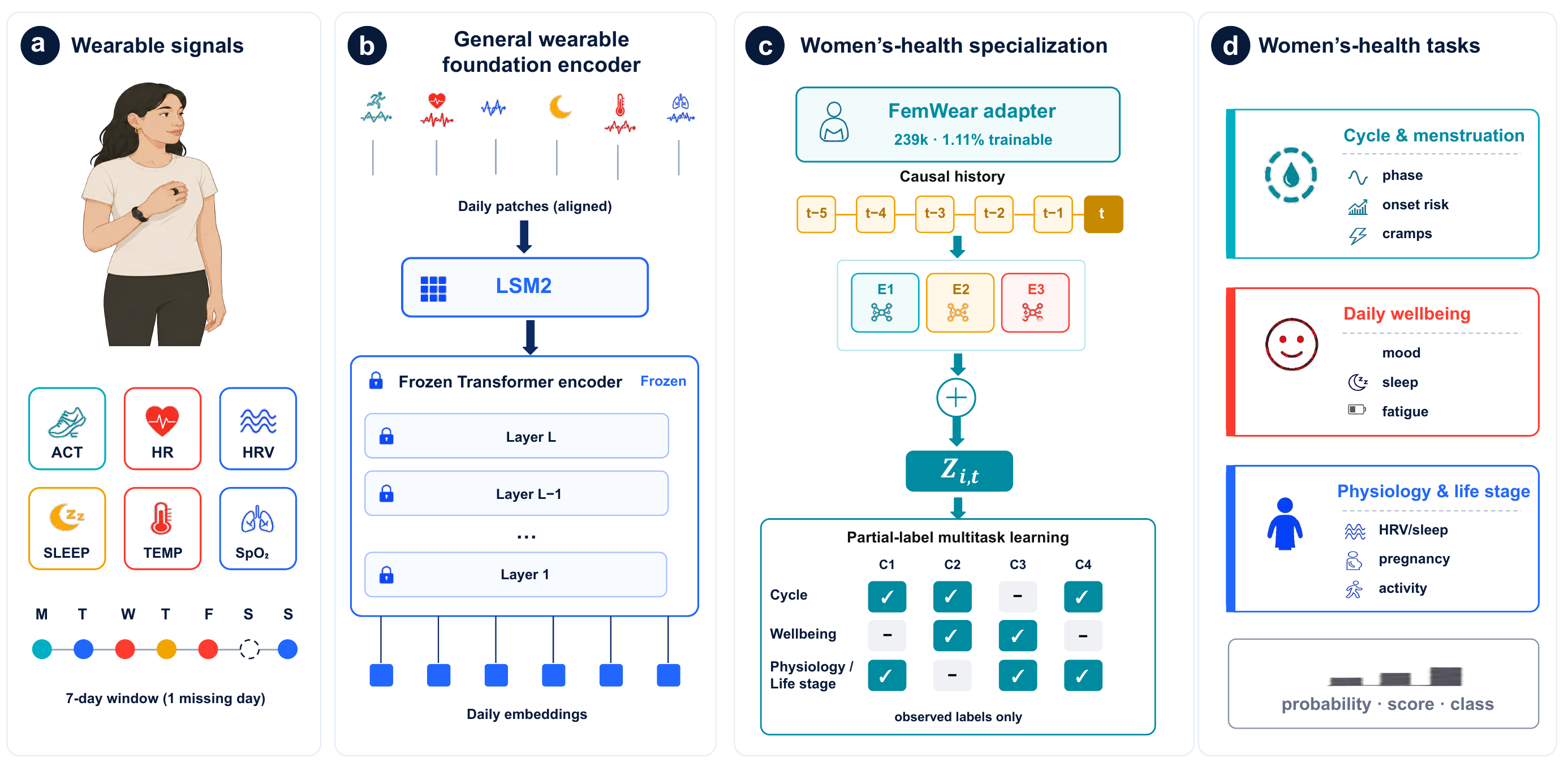}
  \caption{\textbf{Overview of \method.} (a) A seven-day multimodal wearable history is formed from activity, heart rate, heart-rate variability, sleep, temperature, and oxygen signals. (b) A frozen general wearable foundation encoder maps aligned daily patches to daily embeddings. (c) The trainable \method adapter combines causal history with three physiological-regime expert pathways to form a shared women's-health state; partial-label multitask learning uses only labels observed in each cohort. (d) Task-family heads return calibrated probabilities, scores, or classes for cycle and menstrual outcomes, daily wellbeing, and physiology or life-stage outcomes.}
  \label{fig:overview}
\end{figure}

\section{Related Work}

\paragraph{Wearable foundation models.}
Large sensor models show that increasing data, model size, and compute improves imputation and downstream transfer across wearable modalities \citep{narayanswamy2024scaling}. PaPaGei demonstrated open foundation representations for photoplethysmography across cardiovascular, sleep, pregnancy, and wellbeing tasks \citep{pillai2025papagei}. \openmhc provides the open-weight and benchmark substrate used here \citep{schuetz2026openmhc}. \method is a women's wearable foundation model rather than a new from-scratch pretraining effort: it inherits a general wearable representation and repurposes it for women's-health task families.

\paragraph{Foundation-model specialization.}
Domain work commonly distinguishes broad pretraining from parameter-efficient repurposing of a frozen backbone. Gen-P-Tuning adapts a frozen time-series foundation model to multivariate healthcare data \citep{liu2025genptuning}; FORMED repurposes a generic time-series foundation model into a reusable medical time-series classifier and adapts only lightweight task parameters to new datasets \citep{huang2026formed}. \method follows this formulation in the wearable setting. Its foundation-model claim rests on an inherited general wearable encoder, a shared representation across multiple women's-health task families, and lightweight specialization---not on a new large-scale, women-only pretraining run.

\paragraph{Women's wearable health.}
Prior studies relate wearable temperature, pulse, sleep, and activity to menstrual and pregnancy dynamics \citep{lin2024temperature,shilaih2017pulse,ravindra2023pregnancy}. mcPHASES is especially valuable because it synchronizes multimodal Fitbit signals, symptom diaries, glucose, and urine hormone measurements in 42 menstruating participants \citep{lin2025mcphases}. These studies establish that women's-health states leave measurable wearable signatures, but they do not provide a shared wearable representation and adaptation protocol spanning menstrual, mood, sleep, autonomic, activity, and pregnancy tasks.

\paragraph{Parameter-efficient and multitask adaptation.}
Residual adapters and low-rank updates reduce the parameters needed to specialize pretrained models \citep{houlsby2019adapters,hu2022lora}. Multi-gate mixture-of-experts (MMoE) models learn task-specific gates over shared experts \citep{ma2018mmoe}, but heterogeneous objectives may still compete \citep{standley2020tasks}. We therefore compare \method against both simple shared models and MMoE under matched trainable capacity.

\section{Method}

\subsection{Problem formulation}

For participant $i$ and day $t$, let $x_{i,t}$ denote the available wearable channels and $m_{i,t}$ their observation mask. Cohort $c$ exposes labels only for a subset $\mathcal{T}_c$ of the global task registry $\mathcal{T}$. The encoder produces a daily representation
\begin{equation}
  z_{i,t}=E_{\theta_0,\phi}(x_{i,t},m_{i,t}),
\end{equation}
where $\theta_0$ denotes frozen \openmhc parameters and $\phi$ denotes trainable specialization parameters. A causal history model maps $z_{i,1:t}$ to a participant state $h_{i,t}$, and each observed task $k$ has a task-family head $f_k(h_{i,t})$. Labels absent from a cohort are never converted to negatives and do not enter the objective.

\subsection{Parameter-efficient women's-health specialization}

The encoder reuses the pretrained \openmhc LSM2 patch projection and Transformer \citep{vaswani2017attention}. Sensor channels are aligned through descriptors that encode sensor identity, modality, unit, body location, and sampling characteristics, avoiding reliance on device-specific column positions. The pretrained projection and Transformer weights are frozen. Rank-32 bottleneck adapters are inserted into the final two Transformer blocks. For hidden tokens $H$, each zero-initialized residual update is
\begin{equation}
  \operatorname{Adapter}(H)=H+W_{\mathrm{up}}\,\operatorname{GELU}\!\left(W_{\mathrm{down}}\operatorname{LN}(H)\right),
\end{equation}
where $W_{\mathrm{down}}\in\mathbb{R}^{32\times d}$ and $W_{\mathrm{up}}\in\mathbb{R}^{d\times 32}$. A three-expert physiological-regime adapter bank additionally computes a soft mixture of rank-32 residual updates from the pooled daily representation. Together with semantic sensor alignment, the trainable encoder contains 239,236 parameters out of 21,537,796 (1.1108\%).

\subsection{Causal multitask heads}

The downstream model observes only days up to prediction time. A gated recurrent unit \citep{cho2014gru} summarizes the daily sequence. Task-family heads cover menstrual, sleep/recovery, affect/stress, autonomic, activity, cardiometabolic, life-stage, and contextual outcomes. For a batch from cohort $c$, the partial-label loss is
\begin{equation}
  \mathcal{L}_c=\frac{1}{|\mathcal{D}_c|}\sum_{d\in\mathcal{D}_c}
  \frac{1}{|\mathcal{T}_{c,d}|}\sum_{k\in\mathcal{T}_{c,d}}
  \ell_k\!\left(f_k(h),y_k\right),
\end{equation}
where $\mathcal{D}_c$ is the set of active health domains and $\mathcal{T}_{c,d}$ contains observed tasks in domain $d$. Averaging first within domains prevents cohorts with more registered labels from dominating solely through task count.

Menstrual onset is represented by a coherent three-bin distribution over onset within 24 hours, onset between 24 and 72 hours, and onset later than 72 hours. If the corresponding probabilities are $(p_0,p_1,p_2)$, then $P_{24}=p_0$ and $P_{72}=p_0+p_1$, which enforces $P_{24}\leq P_{72}$ for every example. Scalar temperature scaling is fitted using training participants only \citep{guo2017calibration}.

\subsection{Women's-health continual specialization}

The locked women's-health adapter is continually trained for 1,000 steps on the training partitions of five women's-health cohorts, totaling 23,222 participant-days. Cohorts are sampled proportional to the square root of their training-day counts. The objective combines 15\% masked-patch reconstruction, consistency between two corrupted views, preservation of the locked adapter, and retention of the native \openmhc representation. Sensor-channel dropping is applied with probability 0.35. The final-step checkpoint is used; validation early stopping is disabled.

\section{Experimental Setup}

\subsection{Cohorts and tasks}

Table~\ref{tab:cohorts} summarizes the six evaluated sources. OpenMHC-XS is a general ability-retention cohort and is not treated as women-only. The remaining sources contribute menstrual, affective, sleep/HRV, and pregnancy supervision. mcPHASES contains 42 menstruating participants observed across two three-month collection periods \citep{lin2025mcphases}. The pregnancy activity data derive from the gestational-age clock release \citep{ravindra2023pregnancy}. Within every source, participants are disjoint across training, validation, and test partitions. The joint evaluation generates 69 task outputs; 63 have finite and comparable primary metrics, including 33 from women's-health cohorts.

\begin{table}[t]
\centering
\begin{threeparttable}
\caption{Cohorts used to specialize and evaluate \method.}
\label{tab:cohorts}
\small
\begin{tabularx}{\textwidth}{@{}YrrY@{}}
\toprule
Source & Participants & Days/windows & Main wearable signals \\
\midrule
OpenMHC-XS retention & 287 & 53,313 days & activity, heart rate, sleep \\
mcPHASES & 42 & 5,546 encoded days & steps, heart rate, HRV, temperature, oxygen variation, sleep \\
DEPRESS Fitbit, female subset & 66 & 5,728 days & steps, heart rate, sleep \\
inPHRsym, female subset & 44 & 4,882 days & steps, heart rate, sleep \\
Wearable HRV/sleep, female subset & 25 & 508 days & steps, heart rate, HRV, light \\
Pregnancy GA clock & 1,230 & 2,463 seven-day windows & wrist activity, light \\
\bottomrule
\end{tabularx}
\begin{tablenotes}[flushleft]\footnotesize
\item HRV denotes heart-rate variability. Participant counts describe the processed release used by this study; windowed pregnancy data are not treated as independent participants.
\end{tablenotes}
\end{threeparttable}
\end{table}

\subsection{Baselines and controls}

The upstream \openmhc representation is the primary transfer reference and is evaluated with the same downstream protocol. The capacity-matched architecture experiment compares four models with approximately 1.97--1.99M trainable downstream parameters: a latest-observed-day multilayer perceptron, a shared causal GRU, an eight-expert MMoE, and the \method dual-path temporal model. All arms use identical 768-dimensional daily inputs, task registries, participant splits, cohort draws, optimization schedules, seeds, and fixed final checkpoints. This isolates the contribution of temporal and routing structure from parameter-count and checkpoint-selection advantages.

\subsection{Metrics and statistical analysis}

Macro-F1 is primary for multiclass phase recognition. Area under the precision--recall curve (AUPRC) is primary for rare binary events because class prevalence is low \citep{saito2015pr}. Mean absolute error (MAE) is primary for symptom severity and continuous targets. AUROC, Brier score, expected calibration error (ECE), RMSE, and rank or linear correlation are secondary as appropriate. Every reported relative improvement is oriented so that positive values favor \method; decreases in MAE, Brier score, and ECE therefore appear as positive improvements.

Random-seed experiments use seeds 17, 42, and 73. Participant-cluster bootstrap resamples complete participants with replacement for 2,000 replicates, preserving within-participant longitudinal dependence. Holm correction is applied within each prespecified task family \citep{holm1979procedure}. The participant, not the participant-day, is the independent evaluation unit. Test labels are not used for model or checkpoint selection.

\subsection{Implementation}

The encoder uses the public \openmhc LSM2 checkpoint. Adapter rank is 32, the final two Transformer blocks receive internal adapters, and downstream joint models contain approximately 1.98M trainable parameters. Encoder continual pretraining and downstream capacity-matched runs use 1,000 optimization steps. The capacity comparison uses batch size 16; continual pretraining uses batch size 8, AdamW with learning rate $10^{-4}$, weight decay 0.01, and gradient clipping at 1.0. Experiments ran on a single NVIDIA GeForce RTX 5090 D GPU. Recorded 1,000-step downstream runs required approximately 42--132 seconds depending on architecture.

\section{Results}

\subsection{FemWear supports selected menstrual-health endpoints}

On the fixed participant split, \method improved four of the six prespecified menstrual-health outputs in every seed (Table~\ref{tab:fixed}). Cycle-phase macro-F1 increased from 0.4162 to $0.4500\pm0.0397$ (+8.15\%). Cramps, mood symptoms, and sleep-problem MAE decreased by 9.32\%, 5.80\%, and 9.43\%, respectively. The 72-hour onset AUPRC improved by 1.81\% in two of three seeds. In contrast, 24-hour onset AUPRC decreased by 3.40\%, so specialization did not uniformly help even closely related horizons.

\begin{table}[t]
\centering
\begin{threeparttable}
\caption{Fixed participant-split menstrual-health results over three seeds.}
\label{tab:fixed}
\small
\begin{tabularx}{\textwidth}{@{}Ylrrr@{}}
\toprule
Task & Metric & OpenMHC & FemWear, mean $\pm$ SD & Rel. change \\
\midrule
Cycle phase & Macro-F1 $\uparrow$ & 0.4162 & \better{0.4500 $\pm$ 0.0397} & \better{+8.15\%} \\
Onset within 24 h & AUPRC $\uparrow$ & \better{0.0843} & 0.0813 $\pm$ 0.0138 & $-$3.40\% \\
Onset within 72 h & AUPRC $\uparrow$ & 0.2002 & \better{0.2027 $\pm$ 0.0345} & +1.81\% \\
Cramps & MAE $\downarrow$ & 1.0268 & \better{0.9306 $\pm$ 0.0155} & \better{+9.32\%} \\
Mood symptoms & MAE $\downarrow$ & 1.4301 & \better{1.3465 $\pm$ 0.0252} & \better{+5.80\%} \\
Sleep problems & MAE $\downarrow$ & 1.4405 & \better{1.3048 $\pm$ 0.0986} & \better{+9.43\%} \\
\bottomrule
\end{tabularx}
\begin{tablenotes}[flushleft]\footnotesize
\item Relative changes are direction-normalized; positive values favor FemWear. This development split contains six validation participants and is therefore interpreted together with the nested audit.
\end{tablenotes}
\end{threeparttable}
\end{table}

\subsection{Participant-level audit localizes the observed transfer}

\begin{figure}[t]
  \centering
  \includegraphics[width=\linewidth]{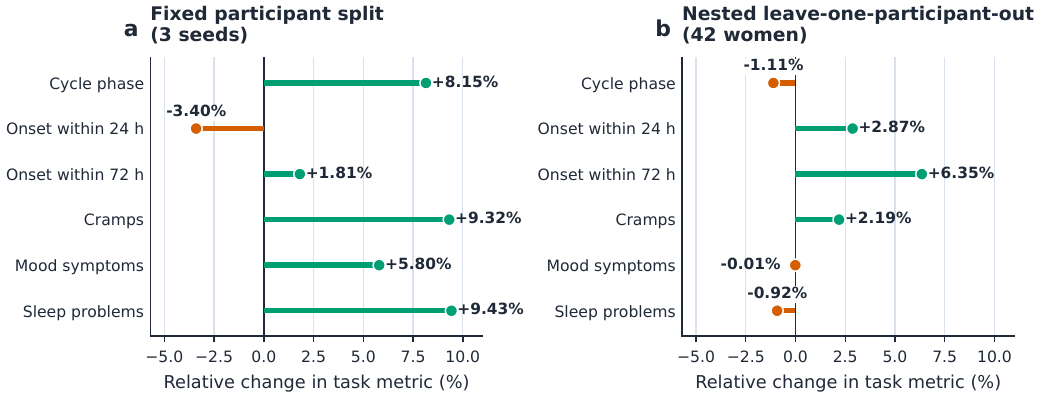}
  \caption{\textbf{Menstrual transfer depends on the evaluation protocol.} Relative changes are oriented so that positive values favor \method. (a) Fixed participant split over three seeds. (b) Single-seed nested leave-one-participant-out evaluation across all 42 mcPHASES participants. The stricter audit preserves positive onset and cramps trends but attenuates phase, mood, and sleep improvements.}
  \label{fig:menstrual}
\end{figure}

The 42-participant nested leave-one-participant-out audit produced a narrower result (Table~\ref{tab:nested} and Figure~\ref{fig:menstrual}). The strongest positive change was 72-hour onset AUPRC (+6.35\%), followed by 24-hour onset AUPRC (+2.87\%) and cramps MAE (+2.19\%). Cycle phase, mood, and sleep did not improve. No endpoint had a strictly positive confidence interval after the prespecified multiple-comparison procedure. The fixed-split gains therefore indicate promising task-specific transfer, while the nested audit marks the current evidential boundary.

\begin{table}[t]
\centering
\caption{Nested leave-one-participant-out audit across 42 mcPHASES participants.}
\label{tab:nested}
\small
\begin{tabularx}{\textwidth}{@{}Ylrrr@{}}
\toprule
Task & Metric & OpenMHC & FemWear & Rel. change \\
\midrule
Cycle phase & Macro-F1 $\uparrow$ & \better{0.2988} & 0.2955 & $-$1.11\% \\
Onset within 24 h & AUPRC $\uparrow$ & 0.0389 & \better{0.0401} & +2.87\% \\
Onset within 72 h & AUPRC $\uparrow$ & 0.1152 & \better{0.1225} & \better{+6.35\%} \\
Cramps & MAE $\downarrow$ & 1.4547 & \better{1.4229} & +2.19\% \\
Mood symptoms & MAE $\downarrow$ & \better{1.5927} & 1.5929 & $-$0.01\% \\
Sleep problems & MAE $\downarrow$ & \better{1.3908} & 1.4036 & $-$0.92\% \\
\bottomrule
\end{tabularx}
\end{table}

\subsection{Temporal modeling contributes, but no architecture is uniformly best}

The capacity-matched experiment removed validation checkpoint selection and constrained every model to approximately 1.98M parameters. \method achieved a validation loss of $0.7691\pm0.0027$, improving over the latest-day MLP by 0.57\%. It beat the MLP on 43 of 63 tasks and 23 of 33 female tasks. The shared GRU and MMoE obtained losses of $0.7681\pm0.0058$ and $0.7689\pm0.0093$, respectively, and \method failed the preregistered majority-task criteria against both (Table~\ref{tab:capacity}). No architecture pair produced a Holm-significant systematic advantage across at least two seeds. Figure~\ref{fig:capacity}a shows that temporal models are tightly grouped and that the latest-day MLP has greater across-seed dispersion.

\begin{table}[t]
\centering
\caption{Fixed-step capacity-matched architecture comparison.}
\label{tab:capacity}
\small
\begin{tabularx}{\textwidth}{@{}YrrrY@{}}
\toprule
Model & Trainable params & Validation loss & Utility & FemWear decision \\
\midrule
FemWear dual path & 1,983,696 & $0.7691\pm0.0027$ & 0.5422 & Reference \\
Latest-day shared MLP & 1,983,530 & $0.7736\pm0.0085$ & 0.3999 & Passed: 43/63; 23/33 female \\
Shared GRU & 1,985,683 & $\mathbf{0.7681\pm0.0058}$ & 0.5617 & Failed: 31/63; 16/33 female \\
MMoE, eight experts & 1,967,366 & $0.7689\pm0.0093$ & \textbf{0.5663} & Failed: 35/63; 17/33 female \\
\bottomrule
\end{tabularx}
\end{table}

\subsection{Continual specialization produces domain-dependent transfer}

\begin{figure}[t]
  \centering
  \includegraphics[width=\linewidth]{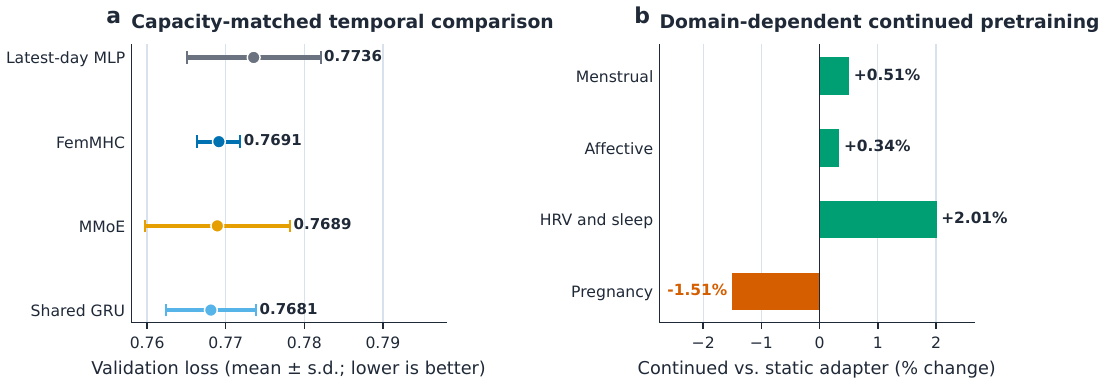}
  \caption{\textbf{Capacity control and transfer boundary.} (a) Mean final validation loss with standard deviation over seeds 17, 42, and 73 for parameter-matched downstream architectures. Lower is better. (b) Relative domain loss change after women's-health multi-cohort continual specialization compared with the locked static adapter. Positive values favor continued specialization.}
  \label{fig:capacity}
\end{figure}

Women's-health continual specialization reduced the static-adapter validation loss by 0.59\%, from 0.7610 to 0.7565, but remained 0.36\% worse than the original \openmhc representation at 0.7539. Relative to the static adapter, it improved the HRV/sleep domain by 2.01\%, menstrual tasks by 0.51\%, and affective tasks by 0.34\%, while pregnancy loss worsened by 1.51\% (Figure~\ref{fig:capacity}b). It won only 19 of 63 tasks and 11 of 33 women's-health tasks against the static adapter, so additional seeds were not run under the preregistered stopping rule. The experiment shows partial repair of the static adapter.

\subsection{Onset probabilities are coherent and calibratable}

Train-only temperature scaling substantially improved probability reliability without changing ranking. For 24-hour onset, AUPRC and AUROC were 0.0801 and 0.7268; calibrated Brier score and ECE were 0.0300 and 0.0097. For 72-hour onset, AUPRC and AUROC were 0.2031 and 0.7343; calibrated Brier score and ECE were 0.0825 and 0.0305. Calibration reduced ECE by 88.2\% and 84.2\% for the two horizons and reduced Brier score by 19.8\% and 30.7\% (Figure~\ref{fig:reliability}a). The shared three-bin head produced zero $P_{24}>P_{72}$ violations in every seed. These results establish coherent, calibratable probability outputs---not product-level discrimination.

\subsection{Recent history is more important than random completeness}

\begin{figure}[t]
  \centering
  \includegraphics[width=\linewidth]{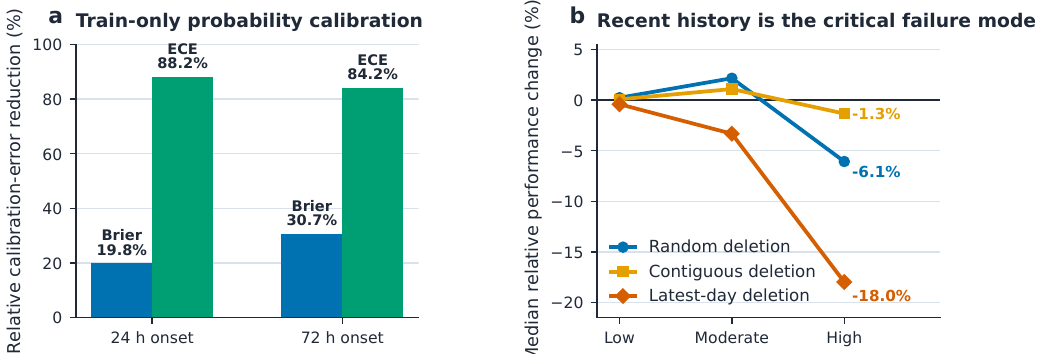}
  \caption{\textbf{Reliability analyses.} (a) Relative reductions in Brier score and expected calibration error (ECE) after scalar temperature scaling fitted on training participants only. (b) Median task performance under increasingly severe random, contiguous, and latest-day deletions. ``Low'', ``moderate'', and ``high'' correspond to 10/25/40\% random deletion and 1/3/7-day deletion for the other conditions.}
  \label{fig:reliability}
\end{figure}

The missing-history audit showed modest changes under 10--25\% random deletion and one-to-three-day contiguous deletion (Table~\ref{tab:missingness}). At 40\% random deletion, median performance decreased by 6.07\%. Removing the latest seven observed days was substantially more damaging, decreasing median performance by 17.99\%, compared with 1.33\% for a generic seven-day contiguous gap (Figure~\ref{fig:reliability}b). The model therefore tolerates moderate nonspecific missingness but depends on recent observations.

\subsection{Few-label behavior}

At 1\%, 5\%, 10\%, 25\%, and 100\% of training labels, \method won 2, 3, 3, 2, and 3 of six menstrual tasks (Table~\ref{tab:label-efficiency}). Median relative changes were $-1.28\%$, $-0.27\%$, $-1.59\%$, $-2.10\%$, and $-0.35\%$, respectively.

\section{Discussion}

The main finding is that \method is viable as a parameter-efficient women's wearable foundation model, with benefits that are endpoint-dependent. With only 1.11\% of encoder parameters trainable, \method produced clear fixed-split gains for cycle phase, cramps, mood, and sleep symptoms. The stricter nested participant audit localized the remaining positive trend to menstrual onset and cramps. This distinction matters because participant-days are abundant relative to independent participants: 5,546 mcPHASES days do not replace the uncertainty imposed by 42 individuals.

The capacity controls further specify the algorithmic contribution. Temporal aggregation improved upon a model that uses only the latest day, but the proposed dual-path design did not establish a stable advantage over an equally sized shared GRU or MMoE. This is consistent with multitask learning theory and empirical work showing that sharing can help some task combinations and harm others \citep{standley2020tasks}. Women's health is not one homogeneous domain: continued specialization improved HRV/sleep more than menstrual or affective tasks and degraded pregnancy transfer. A useful specialized foundation model must therefore control negative transfer rather than merely add domain data.

Calibration gives a complementary result. The onset head enforces logically nested horizons, while train-only temperature scaling substantially reduces Brier score and ECE. Good calibration does not compensate for low AUPRC, but it makes the meaning of an output probability more defensible. In practical evaluation, discrimination, calibration, and temporal consistency should be reported separately rather than collapsed into one accuracy number.

Several limitations remain. The core menstrual cohort contains 42 participants, and the nested audit is internal cross-validation rather than independent external validation. The fixed development split contains only six validation participants. The women's HRV/sleep cohort contains 25 participants, and target definitions vary across sources. Daily cached representations prevent the joint objective from adapting raw minute-level sensor features end to end. Missingness experiments remove daily history but do not simulate channel-specific sensor noise or motion artifacts.

These limitations set the agenda for the next experiment. The immediate methodological target is a negative-transfer-aware adapter that preserves the strong general representation while activating women's-health specialization only when task and cohort evidence support it. The immediate evidential target is repeated nested evaluation and a participant-disjoint menstrual cohort collected under a prespecified protocol.

\section{Conclusion}

\method is a women's wearable foundation model, built by training 239,236 encoder parameters on top of a general open wearable foundation backbone. It improves selected menstrual, symptom, mood, and sleep endpoints on a fixed split; preserves positive onset and cramps trends under stricter participant-level validation; and produces coherent, calibratable onset probabilities. The central contribution is a reproducible specialized-model formulation and evaluation suite for women's-health wearables, together with a precise account of where its parameter-efficient transfer is strongest.

\section*{Data and Code Availability}

Code, configurations, processed-data interfaces, and aggregate experimental artifacts are available at \url{https://github.com/YifanWang-China/FemMHC}. Raw participant-level datasets, upstream \openmhc weights, training checkpoints, and individual-level predictions are excluded from version control and remain subject to their source licenses and data-use agreements.

\bibliographystyle{unsrtnat}
\bibliography{references}

\appendix
\section{Evaluation Protocols and Metric Definitions}

For multiclass cycle phase, the primary metric is macro-F1 so that all phases contribute equally. For rare binary onset and affect events, AUPRC is primary and AUROC is secondary. For symptom severity and continuous physiology, MAE is primary; RMSE and Pearson or Spearman correlation are secondary. Brier score and ECE measure probability quality. Unless otherwise stated, metrics are calculated over held-out observations and uncertainty is clustered by participant. The fixed split uses a development partition with six validation participants; the nested audit uses leave-one-participant-out cross-validation over all 42 mcPHASES participants. Seeds 17, 42, and 73 are used for all multi-seed results, and multiplicity correction follows Holm's procedure within each prespecified task family.

\section{Implementation and Reproducibility}

All released aggregate experiments record configuration, seed, fixed-step budget, parameter counts, and evaluation artifacts. The locked capacity experiment uses seeds 17, 42, and 73 and participant-disjoint data splits. The adapter specialization adds 239,236 trainable encoder parameters to the frozen 21,537,796-parameter \openmhc LSM2 encoder, and the downstream joint model contains approximately 1.98M trainable parameters. All capacity-matched arms share identical inputs, task registries, splits, cohort draws, optimization schedules, and fixed final checkpoints, so the only free variables are architecture and routing. The manuscript source-data directory contains the exact values used to generate every numerical figure.

\section{Additional Results}

\subsection{Label-efficiency audit}

\begin{table}[h]
\centering
\caption{Equal-width six-task label-efficiency experiment.}
\label{tab:label-efficiency}
\small
\begin{tabular}{@{}rrrrr@{}}
\toprule
Labels & Task wins & Task losses & Median change & Interquartile range \\
\midrule
1\% & 2 & 4 & $-1.28\%$ & [$-2.59$, $+0.97$]\% \\
5\% & 3 & 3 & $-0.27\%$ & [$-4.35$, $+2.89$]\% \\
10\% & 3 & 3 & $-1.59\%$ & [$-3.99$, $+0.45$]\% \\
25\% & 2 & 4 & $-2.10\%$ & [$-4.05$, $+0.94$]\% \\
100\% & 3 & 3 & $-0.35\%$ & [$-2.59$, $+2.95$]\% \\
\bottomrule
\end{tabular}
\end{table}

\subsection{Missing-history conditions}

\begin{table}[h]
\centering
\caption{Median relative task change under deterministic history deletion.}
\label{tab:missingness}
\small
\begin{tabular}{@{}lrrr@{}}
\toprule
Deletion family & Low & Moderate & High \\
\midrule
Random observed days & +0.24\% (10\%) & +2.17\% (25\%) & $-6.07\%$ (40\%) \\
Contiguous block & +0.08\% (1 day) & +1.10\% (3 days) & $-1.33\%$ (7 days) \\
Latest observed days & $-0.40\%$ (1 day) & $-3.32\%$ (3 days) & $-17.99\%$ (7 days) \\
\bottomrule
\end{tabular}
\end{table}

\section{Ethics and Societal Impact}

This study performs secondary analysis of de-identified public or access-controlled research datasets. No model output in this work is evaluated as a diagnosis or treatment recommendation. A women's-health specialization can surface patterns that general wearable models overlook, but it also concentrates risk: menstrual and pregnancy inferences touch sensitive, sometimes stigmatized outcomes, and a poorly calibrated or weakly discriminative predictor could mislead users. We therefore foreground the negative and neutral results alongside the positive ones.

\end{document}